\PassOptionsToPackage{hyphens}{url}
\documentclass[10pt]{wlscirep}
\usepackage[T1]{fontenc}
\usepackage{textcomp}
\usepackage{xurl}
\title{WaveFlow-UIE: A Single-Pipeline Wavelet-Domain Flow Model for Efficient Underwater Image Restoration}

\author[1]{Abdulaziz Alfaraj}
\author[1]{Mohammed Al Naser}
\author[1]{Hassan Al Nasser}
\author[1,2,*]{Muzammil Behzad}
\affil[1]{King Fahd University of Petroleum \& Minerals, Dhahran 31261, Saudi Arabia}
\affil[2]{SDAIA-KFUPM Joint Research Center for Artificial Intelligence, Dhahran 31261, Saudi Arabia}
\affil[*]{muzammil.behzad@kfupm.edu.sa}


\begin{abstract}
Underwater image restoration remains challenging because color distortion, scattering, and detail loss affect low- and high-frequency image components differently. Recent diffusion-based methods have shown strong restoration quality, but often rely on multi-stage pipelines and costly iterative refinement. We present WaveFlow-UIE, a single-pipeline wavelet-domain flow model that directly learns to transform degraded underwater wavelet representations into clean ones. Our method applies a discrete wavelet transform, estimates lightweight physics-inspired degradation cues including transmission and ambient light, and conditions a velocity network trained with flow matching in wavelet space. The training objective combines flow-matching, frequency-weighted reconstruction, perceptual, color-consistency, and physics-consistency losses. Across UIEB, LSUI, UFO-120, and EUVP, WaveFlow-UIE achieves a strong quality-efficiency tradeoff, with especially competitive results on perceptual metrics. Relative to a strong wavelet-based diffusion baseline, the proposed model substantially reduces inference cost while maintaining or improving restoration quality on several paired benchmarks. We show that reducing the inference process from five flow steps to a single step preserves nearly all of the restoration quality while providing an additional speedup of about $4\times$ on UIEB at $256\times256$ resolution. This single-step result suggests that a single-pass wavelet-domain model is an effective alternative to heavier diffusion-style refinement.
\end{abstract}

\begin{document}

\flushbottom
\maketitle
\thispagestyle{empty}

\section*{Introduction}

Underwater image restoration is challenging because light absorption and scattering cause wavelength-dependent color distortion, haze, low contrast, and loss of fine detail~\cite{jaffe1990, akkaynak2018}. These degradations reduce both visual quality and the reliability of downstream perception systems used in marine inspection, robotic navigation, and scientific exploration~\cite{islam2020fast}. Recent learning-based methods have improved underwater enhancement substantially~\cite{li2020uieb, li2020uwcnn, li2021ucolor, peng2023ushape}, but many still operate purely in the spatial domain and do not explicitly model the different behavior of low- and high-frequency image components.

WF-Diff~\cite{zhao2024wavelet} is a strong frequency-domain restoration framework that addresses this issue by decomposing images into wavelet components and applying a two-stage pipeline: a coarse restoration network followed by diffusion-based~\cite{ho2020ddpm} frequency residual refinement. While effective, this design increases architectural complexity and inference cost because the refinement stage remains iterative and separate from the main restoration network.

We propose WaveFlow-UIE, a single-pipeline wavelet-domain flow model for underwater image restoration. Instead of predicting a coarse enhanced image and then refining residuals with diffusion, WaveFlow-UIE directly learns a restoration flow~\cite{lipman2023flow, liu2023rectified} from degraded wavelet representations toward clean wavelet representations. To improve physical plausibility, the model also includes a lightweight physics prior branch that estimates transmission and ambient-light cues~\cite{he2011dcp, drews2013udcp} and uses them to condition the restoration process. Training combines flow matching, frequency-weighted reconstruction, perceptual supervision~\cite{zhang2018lpips}, Lab-space color consistency, and physics consistency.

The main empirical result is a favorable quality-efficiency tradeoff relative to the released WF-Diff baseline. In addition, a single-step inference variant preserves nearly all of the full model quality while yielding an additional speedup of about $4\times$ on UIEB at $256\times256$ resolution, making it one of the strongest practical outcomes of the paper. Consequently, the salient features of our proposed method are as follows:
\begin{itemize}
    \item We reformulate wavelet-domain underwater image restoration as a single-pipeline conditional flow problem.
    \item We introduce physics-prior conditioning through transmission and ambient-light estimation.
    \item We combine frequency-aware, perceptual, and color-consistency supervision and analyze the contribution of each term in a controlled ablation.
    \item We show a strong speed-quality tradeoff relative to the released WF-Diff baseline, including a strong single-step inference result.
    \item We show that reducing inference from five flow steps to one preserves nearly all restoration quality while yielding about a $4\times$ speedup on UIEB at $256\times256$ resolution.
\end{itemize}

\section*{Related work}

\subsection*{Underwater image restoration}
Underwater enhancement methods aim to recover visually clear images from observations degraded by absorption and scattering~\cite{jaffe1990, akkaynak2018}. Classical prior-based approaches, such as dark-channel-based transmission estimation~\cite{he2011dcp, drews2013udcp} and restoration based on image blurriness and light absorption~\cite{peng2017ibla}, are interpretable but often brittle under varying water conditions. Learning-based methods such as Water-Net~\cite{li2020uieb}, UWCNN~\cite{li2020uwcnn}, UIE-WD~\cite{ma2022uiewd}, and U-shape~\cite{peng2023ushape}, learn direct mappings from degraded images to enhanced outputs and generally produce stronger restoration quality, but many operate mainly in RGB image space. Other designs combine multiple color spaces, as in Ucolor~\cite{li2021ucolor} and UIEC$^2$-Net~\cite{wang2021uiec2net}, use adversarial training, as in FUnIE-GAN~\cite{islam2020fast}, or apply conditional diffusion models, as in DiffWater~\cite{guan2024diffwater}.

\subsection*{Frequency-domain restoration}
Frequency-domain representations, such as the multiresolution wavelet decomposition~\cite{mallat1989}, are useful in low-level vision because low-frequency components mainly capture color, illumination, and global structure, while high-frequency components capture edges, texture, and fine detail. In underwater enhancement, UIE-WD~\cite{ma2022uiewd} decomposes the input with the discrete wavelet transform and processes the low-frequency structure and high-frequency detail components in two separate streams, and WF-Diff~\cite{zhao2024wavelet} showed that wavelet and Fourier information can be effectively combined for underwater restoration. Our method keeps the wavelet-domain motivation but replaces the two-stage coarse-plus-diffusion design with a single flow-based pipeline.

\subsection*{Diffusion and flow-based restoration}
Diffusion models~\cite{ho2020ddpm} have achieved strong results in image generation and restoration, including underwater enhancement~\cite{zhao2024wavelet, guan2024diffwater}, but their iterative sampling process can be expensive. Flow matching~\cite{lipman2023flow} and rectified flow~\cite{liu2023rectified} methods provide an alternative by learning a direct transformation field between source and target representations. WaveFlow-UIE adopts this idea in wavelet space, allowing the model to learn a direct restoration flow without a separate diffusion refinement stage.

\section*{Methods}

\subsection*{Overview}

\begin{figure}[!t]
\centering
\includegraphics[width=\linewidth]{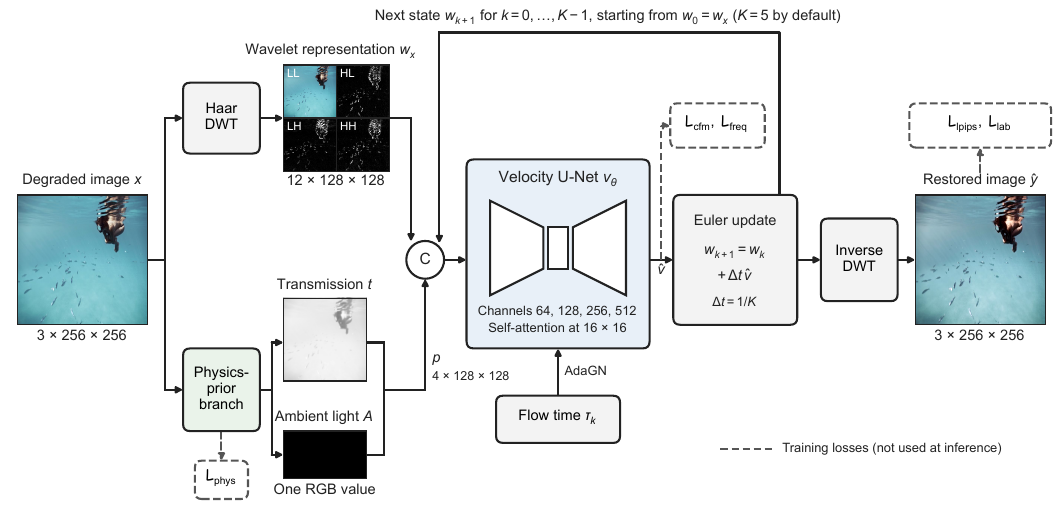}
\caption{\textbf{WaveFlow-UIE architecture.} A degraded underwater image is decomposed with a Haar discrete wavelet transform (DWT), conditioned by a lightweight physics-prior branch that predicts transmission and ambient-light cues, and restored by a velocity U-Net trained with flow matching in wavelet space. The enhanced image is reconstructed with the inverse wavelet transform after iterative flow updates.}
\label{fig:model-architecture}
\end{figure}

WaveFlow-UIE restores a degraded underwater image by learning a transformation from its degraded wavelet representation to the corresponding clean wavelet representation. Given a degraded image, the model first applies a discrete wavelet transform to separate low- and high-frequency components. A lightweight physics prior branch estimates degradation cues, including transmission and ambient light, and these cues are used to condition a velocity network trained with flow matching. The final restored image is obtained through inverse wavelet reconstruction (Fig.~\ref{fig:model-architecture}).

\subsection*{Wavelet-domain representation}

Underwater degradation affects different visual components differently: color distortion and illumination shifts mainly appear in low-frequency components, while texture loss and edge degradation mainly appear in high-frequency components. We therefore operate in wavelet space instead of directly only in RGB space.

Given a degraded underwater image $x \in \mathbb{R}^{3 \times H \times W}$, we apply the Haar discrete wavelet transform~\cite{mallat1989}:
\begin{equation}
    w_x = \mathrm{DWT}(x),
\end{equation}
where $w_x \in \mathbb{R}^{12 \times \frac{H}{2} \times \frac{W}{2}}$. Similarly, for the clean reference image $y$,
\begin{equation}
    w_y = \mathrm{DWT}(y).
\end{equation}
This representation preserves the information needed for exact reconstruction while separating global structure from fine detail.

\subsection*{Physics-prior conditioning}

Underwater image degradation is strongly tied to light propagation in water. Inspired by simplified underwater formation models~\cite{jaffe1990, drews2013udcp}, we add a lightweight branch that predicts a per-pixel transmission map $t$ and a single global ambient light $A \in [0,1]^3$ (one RGB value per image):
\begin{equation}
    x \approx y \cdot t + A \cdot (1 - t).
\end{equation}
The transmission map is downsampled to wavelet resolution, the ambient light is broadcast to the same spatial size, and the two are concatenated into a four-channel conditioning tensor:
\begin{equation}
    p = [t, A].
\end{equation}
This conditioning is intended to give the restoration model explicit information about haze and color cast; the values it takes in practice are examined in the Discussion.

\subsection*{Wavelet-domain flow matching}

Instead of using a separate enhancement stage followed by diffusion refinement, WaveFlow-UIE directly learns a restoration flow in wavelet space~\cite{lipman2023flow, liu2023rectified}. During training, we sample a time value $\tau \sim \mathcal{U}(0,1)$ and construct an interpolated wavelet state:
\begin{equation}
    w_{\tau} = (1 - \tau) w_x + \tau w_y.
\end{equation}
The target velocity is the direction from the degraded wavelet representation to the clean wavelet representation:
\begin{equation}
    v^\star = w_y - w_x.
\end{equation}
A conditional velocity network predicts this direction:
\begin{equation}
    \hat{v} = v_{\theta}(w_{\tau}, \tau, w_x, p),
\end{equation}
where $w_{\tau}$ is the current wavelet state, $\tau$ is the flow time, $w_x$ is the degraded condition, and $p$ is the physics-prior condition.

The predicted clean wavelet representation is:
\begin{equation}
    \hat{w}_y = w_x + \hat{v},
\end{equation}
and the restored image is reconstructed with the inverse wavelet transform:
\begin{equation}
    \hat{y} = \mathrm{IDWT}(\hat{w}_y).
\end{equation}

\subsection*{Training objective}

The total training loss combines several complementary terms:
\begin{equation}
\begin{split}
    \mathcal{L}_{\mathrm{total}} ={}&
    \lambda_{\mathrm{cfm}} \mathcal{L}_{\mathrm{cfm}}
    + \lambda_{\mathrm{freq}} \mathcal{L}_{\mathrm{freq}}
    + \lambda_{\mathrm{lpips}} \mathcal{L}_{\mathrm{lpips}} \\
    &+ \lambda_{\mathrm{lab}} \mathcal{L}_{\mathrm{lab}}
    + \lambda_{\mathrm{phys}} \mathcal{L}_{\mathrm{phys}}.
\end{split}
\end{equation}
Here, $\mathcal{L}_{\mathrm{cfm}}$ is the conditional flow matching loss, $\mathcal{L}_{\mathrm{freq}}$ is a frequency-weighted wavelet reconstruction loss that emphasizes high-frequency detail, $\mathcal{L}_{\mathrm{lpips}}$ is the perceptual loss~\cite{zhang2018lpips}, $\mathcal{L}_{\mathrm{lab}}$ enforces Lab-space color consistency, and $\mathcal{L}_{\mathrm{phys}}$ supervises the physics-prior branch. This combination is intended to preserve both color fidelity and fine detail while keeping the model physically grounded.
With $\Delta v = \hat{v} - v^\star$, the individual terms are
\begin{align}
    \mathcal{L}_{\mathrm{cfm}} &= \mathrm{mean}\left(\Delta v^{2}\right), \\
    \mathcal{L}_{\mathrm{freq}} &= \mathrm{mean}\left|\Delta v_{\mathrm{LL}}\right| + \gamma\,\mathrm{mean}\left|\Delta v_{\mathrm{HF}}\right|, \\
    \mathcal{L}_{\mathrm{lpips}} &= \mathrm{LPIPS}(\hat{y}, y), \\
    \mathcal{L}_{\mathrm{lab}} &= \mathrm{mean}\left|\mathrm{ab}(\hat{y}) - \mathrm{ab}(y)\right|, \\
    \mathcal{L}_{\mathrm{phys}} &= \mathrm{mean}\left(\left(x - y \cdot t - A \cdot (1 - t)\right)^{2}\right),
\end{align}
where the means run over all coefficients or pixels of a batch, $\Delta v_{\mathrm{LL}}$ holds the three low-frequency channels of $\Delta v$, $\Delta v_{\mathrm{HF}}$ holds its nine high-frequency channels, $\gamma = 2$, and $\mathrm{ab}(\cdot)$ denotes the a and b channels of the CIELAB color space. LPIPS uses the AlexNet backbone with images scaled to $[-1, 1]$, and $\hat{y}$ is obtained from $\hat{w}_y = w_x + \hat{v}$ at every sampled $\tau$. The weights are $\lambda_{\mathrm{cfm}} = 1$, $\lambda_{\mathrm{freq}} = 0.5$, $\lambda_{\mathrm{lpips}} = 0.1$, $\lambda_{\mathrm{lab}} = 0.05$, and $\lambda_{\mathrm{phys}} = 0.1$.

\subsection*{Network architecture}

The velocity network $v_\theta$ is a U-Net with a 28-channel input (the 12-channel state, the 12-channel condition $w_x$, and the 4-channel condition $p$) and a 12-channel output. Its four resolution levels have 64, 128, 256, and 512 channels, with two residual blocks per level in the encoder and three per level in the decoder. Residual blocks use group normalization with 32 groups and a dropout rate of $0.1$. The flow time passes through a 64-dimensional sinusoidal embedding and a two-layer perceptron with SiLU activation to give a 256-dimensional embedding, which modulates each residual block through adaptive group normalization (scale and shift). Multi-head self-attention with eight heads is used only at the deepest level ($16\times16$ feature maps for a $256\times256$ input): twice in the encoder, once in the bottleneck, and three times in the decoder. Downsampling uses strided convolutions, and upsampling uses nearest-neighbor interpolation followed by a convolution.

The physics-prior branch consists of three $3\times3$ convolutions with $3 \to 32 \to 64 \to 32$ channels and ReLU activations. A $1\times1$ convolution followed by a sigmoid gives $t$, and global average pooling followed by a linear layer and a sigmoid gives $A$. The model has $63{,}387{,}184$ parameters in total, of which $37{,}988$ belong to the physics-prior branch.

\subsection*{Inference}

At inference time, the model starts from the degraded wavelet representation and updates it with the predicted velocity field:
\begin{equation}
    w_{k+1} = w_k + \Delta t \cdot v_{\theta}(w_k, \tau_k, w_x, p),
\end{equation}
where $k = 0, \dots, K-1$ denotes the inference step, $\tau_k = k\,\Delta t$ is the current flow time, and $\Delta t = 1/K$ is the step size. In the default setting, we use $K=5$ Euler steps. We also evaluate a single-step variant ($K=1$) to study the quality-efficiency tradeoff.

\subsection*{Implementation details}

The model is implemented in PyTorch. We use AdamW with $\beta_1 = 0.9$, $\beta_2 = 0.999$, weight decay $1\times10^{-4}$, an initial learning rate of $2\times10^{-4}$, linear warm-up for the first $5{,}000$ iterations, and cosine decay to $1\times10^{-6}$. Training uses $256\times256$ patches, an effective batch size of $8$ (four images per step with two accumulation steps), mixed precision, and gradient clipping at $1.0$. Training pairs are augmented with random crops, horizontal flips, vertical flips, and transposition (each flip and the transposition with probability $0.5$). Evaluation uses an exponential moving average of the velocity-network weights (decay $0.999$, and $0.99$ during the first $1{,}000$ iterations). If the loss becomes non-finite, training resumes from the most recent checkpoint. The default inference setting uses $K=5$ Euler steps with step size $\Delta t = 0.2$. The main WaveFlow models were trained on a single NVIDIA GeForce RTX 5080 GPU, and all reproduced evaluation runs were measured on a single NVIDIA GeForce RTX 5070 Ti GPU with 16\,GB VRAM. The three main WaveFlow checkpoints were taken at $150$k iterations on UIEB, $240$k iterations on LSUI, and $255$k iterations on the mixed UIEB+LSUI training setup (the cosine schedule was set to $200$k iterations for UIEB and to $2$M iterations for LSUI and the mixed setup). Each is the checkpoint with the highest PSNR in the periodic evaluations made during training. Because no separate validation split was held out, these evaluations used the corresponding test split for the UIEB- and LSUI-trained models, so the reported results of these two checkpoints are likely to be optimistic.

\subsection*{Datasets and evaluation metrics}

We train on the Underwater Image Enhancement Benchmark (UIEB)~\cite{li2020uieb} and the Large-Scale Underwater Image dataset (LSUI)~\cite{peng2023ushape}, and evaluate on UIEB, LSUI, UFO-120~\cite{islam2020sesr}, Enhancing Underwater Visual Perception (EUVP)~\cite{islam2020fast}, U45~\cite{li2019fusion}, and C60~\cite{li2020uieb}. We use the public splits, with 800 training and 90 test pairs for UIEB and 3{,}879 training and 400 test pairs for LSUI. UFO-120 (120 pairs) and EUVP (515 pairs) serve only as paired test sets, and U45 (45 images) and C60 (60 images) as test sets without reference images. For paired benchmarks we report peak signal-to-noise ratio (PSNR$\uparrow$), structural similarity (SSIM$\uparrow$)~\cite{wang2004ssim}, learned perceptual image patch similarity (LPIPS$\downarrow$)~\cite{zhang2018lpips}, and Fr\'echet inception distance (FID$\downarrow$)~\cite{heusel2017gans}. For non-reference benchmarks we additionally report the underwater color image quality evaluation (UCIQE$\uparrow$)~\cite{yang2015uciqe} and the underwater image quality measure (UIQM$\uparrow$)~\cite{panetta2015uiqm}. Arrows indicate whether higher ($\uparrow$) or lower ($\downarrow$) values are better. PSNR and SSIM are computed on the luminance channel after cropping a 2-pixel border. LPIPS uses the AlexNet backbone. FID uses the 2048-dimensional pool3 features of the torchvision InceptionV3 model, with images resized to $299\times299$. Except for FID, values are averaged over the test images. For the standardized $256\times256$ protocol, input and reference images are resized with area interpolation; for the native protocol, images are padded to even dimensions.

\subsection*{Comparison methods}

We compare WaveFlow-UIE against reproduced public checkpoints for WF-Diff~\cite{zhao2024wavelet}, UIE-WD~\cite{ma2022uiewd}, UWCNN~\cite{li2020uwcnn}, Water-Net~\cite{li2020uieb}, U-shape~\cite{peng2023ushape}, and the conditional diffusion model DiffWater~\cite{guan2024diffwater} in the main quantitative benchmark. DiffWater samples with 2{,}000 denoising steps (about 38\,s per $256\times256$ image on our GPU) and was evaluated at $256\times256$ on UIEB, LSUI, and UFO-120. To avoid overclaiming exact protocol reproduction across papers, all reproduced models are evaluated under a unified local benchmark protocol with consistent preprocessing, metric computation, and runtime measurement on the same hardware.

\subsection*{Controlled retraining protocol}

The main checkpoints differ in training length and were chosen by periodic evaluation during training (see Implementation details), so component contributions and the role of flow matching were measured with a separate set of runs. These were trained from scratch on the 800 UIEB training pairs with the settings above, except that they used full precision (mixed-precision runs intermittently produced non-finite losses, whereas the full-precision runs did not), a fixed $100$k-iteration cosine schedule, and no checkpoint selection: the weights at the final iteration were evaluated on the $256\times256$ and native protocols. These runs were trained and evaluated on the RTX 5070 Ti GPU.

The full model was trained with seeds 42, 1, and 2 and evaluated with $K=5$. Each ablation variant was trained with seed 42: no physics prior (branch disabled so that $p = 0$, and $\lambda_{\mathrm{phys}} = 0$), no frequency weighting ($\gamma = 1$), no LPIPS loss ($\lambda_{\mathrm{lpips}} = 0$), and no Lab color loss ($\lambda_{\mathrm{lab}} = 0$). Direct regression uses the same network, losses, and data, but always samples $\tau = 0$ during training; it was trained with seeds 42, 1, and 2 and evaluated with $K=1$. Results over seeds are summarized by the mean and the sample standard deviation (s.d., $n = 3$). Because three seeds are few, per-seed values are also reported (Supplementary Table S4), and no hypothesis tests were applied.

\section*{Results}

\subsection*{Quantitative analysis}

Table~\ref{tab:table-a-uieb-lsui-256} shows that WaveFlow-UIE is especially strong on perceptual metrics at standardized resolution. On UIEB, our method achieves the best LPIPS and the second-best PSNR, SSIM, and FID, indicating that it produces visually closer reconstructions while remaining competitive in pixel-level fidelity. On LSUI, WaveFlow-UIE obtains the best SSIM and LPIPS and the second-best PSNR and FID. These results suggest that the proposed wavelet-domain flow formulation is effective at preserving structure and perceptual detail across the two main paired benchmarks.

\begin{table}[!t]
\centering
\caption{\textbf{Quantitative comparison on the public UIEB and LSUI splits at standardized 256$\times$256 resolution.} Best results are highlighted in bold and second-best results are underlined. The Ours row uses WaveFlow-UIEB on UIEB and WaveFlow-LSUI on LSUI. DiffWater likewise uses its released UIEB and LSUI checkpoints. PSNR, peak signal-to-noise ratio (dB); SSIM, structural similarity; LPIPS, learned perceptual image patch similarity; FID, Fr\'echet inception distance; UCIQE, underwater color image quality evaluation; UIQM, underwater image quality measure. Arrows indicate whether higher ($\uparrow$) or lower ($\downarrow$) is better.}\label{tab:table-a-uieb-lsui-256}
\small
\begin{tabular*}{\linewidth}{@{\extracolsep{\fill}}lcccccc}
\toprule
Method & PSNR$\uparrow$ & SSIM$\uparrow$ & LPIPS$\downarrow$ & FID$\downarrow$ & UCIQE$\uparrow$ & UIQM$\uparrow$ \\
\midrule
\multicolumn{7}{l}{\textit{UIEB}} \\
UIE-WD & 14.8806 & 0.7387 & 0.3771 & 92.3536 & 0.5704 & 4.0653 \\
UWCNN & 19.5062 & 0.8482 & 0.3080 & 72.2604 & 0.5451 & 3.8867 \\
Water-Net & 23.6456 & 0.9106 & \underline{0.1623} & 43.7351 & 0.5826 & \underline{4.2446} \\
U-shape & 24.6389 & 0.8479 & 0.1940 & 48.6329 & \underline{0.5857} & 4.2037 \\
WF-Diff & 24.7457 & 0.9150 & 0.1692 & 53.9730 & 0.5847 & 4.1871 \\
DiffWater & \textbf{25.8002} & \textbf{0.9454} & 0.1672 & \textbf{27.9865} & \textbf{0.6044} & \textbf{4.2604} \\
Ours & \underline{24.9818} & \underline{0.9316} & \textbf{0.1390} & \underline{29.7382} & 0.5835 & 4.0450 \\
\midrule
\multicolumn{7}{l}{\textit{LSUI}} \\
UIE-WD & 17.8399 & 0.8051 & 0.3348 & 53.1968 & 0.5642 & \textbf{4.4957} \\
UWCNN & 21.0453 & 0.8636 & 0.3264 & 54.8434 & 0.5560 & 4.0106 \\
Water-Net & 24.9394 & 0.9015 & 0.2227 & 39.2744 & 0.5857 & \underline{4.4053} \\
U-shape & \textbf{29.1284} & \underline{0.9172} & \underline{0.1449} & \textbf{22.9495} & \underline{0.5865} & 4.3093 \\
WF-Diff & 28.2709 & 0.9112 & 0.1506 & 33.2456 & 0.5734 & 4.3176 \\
DiffWater & 22.9086 & 0.8529 & 0.2712 & 37.9388 & \textbf{0.5958} & 4.3802 \\
Ours & \underline{28.4551} & \textbf{0.9246} & \textbf{0.1216} & \underline{25.9970} & 0.5743 & 4.2832 \\
\bottomrule
\end{tabular*}
\end{table}

\begin{figure}[!t]
\centering
\includegraphics[width=\linewidth]{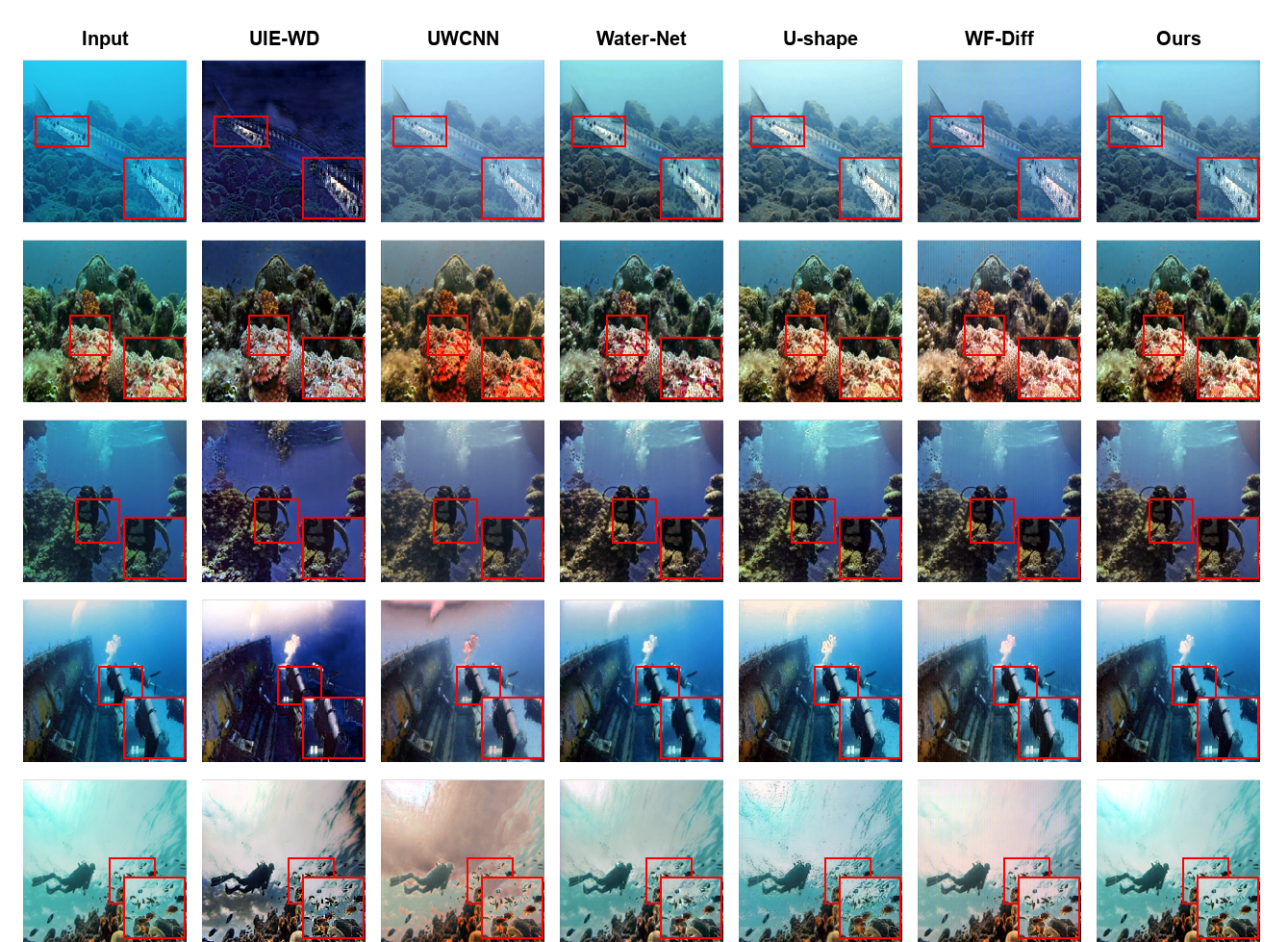}
\caption{\textbf{Representative qualitative comparison on five UIEB test examples.} We compare against UIE-WD, UWCNN, Water-Net, U-shape, and WF-Diff. WaveFlow-UIE typically restores stronger color contrast than UIE-WD and UWCNN while preserving fine structure more consistently than the heavier WF-Diff baseline. Test images are from the UIEB dataset~\cite{li2020uieb}. Each column shows the saved output of the named method, and Ours is the released WaveFlow-UIEB checkpoint with $K=5$. Red boxes mark the region enlarged in the inset.}
\label{fig:qualitative-results}
\end{figure}

Figure~\ref{fig:qualitative-results} provides a qualitative view of the same comparison. Across these examples, WaveFlow-UIE tends to recover more balanced color and clearer local detail than the weaker feed-forward baselines, while remaining visually competitive with the stronger wavelet-based methods. This qualitative behavior is consistent with the quantitative pattern in Table~\ref{tab:table-a-uieb-lsui-256}, where the model is strongest on perceptual fidelity.

Table~\ref{tab:table-b-other-datasets-256} summarizes cross-dataset evaluation on UFO-120, EUVP, U45, and C60 at standardized $256\times256$ resolution. Using the LSUI-trained checkpoint, WaveFlow-UIE generalizes strongly on the paired benchmarks: on UFO-120 it achieves the best SSIM, LPIPS, and FID while remaining second-best in PSNR, indicating strong preservation of structure and perceptual quality under domain shift. On EUVP, our method obtains the best LPIPS, showing that its outputs remain perceptually close to the reference images, although U-shape and WF-Diff are stronger on PSNR, SSIM, and FID. On the non-reference sets U45 and C60, the mixed-data WaveFlow-BOTH checkpoint remains competitive but does not lead the UIQM/UCIQE rankings, suggesting that WaveFlow generalizes more reliably on paired cross-dataset benchmarks than on fully reference-free challenge images.

\begin{table}[!t]
\centering
\caption{\textbf{Benchmark results on UFO-120, EUVP, U45, and C60 at 256$\times$256.} DiffWater was evaluated on UFO-120 only, with its released LSUI checkpoint. Best results are in bold and second-best results are underlined. Dashes mark combinations with no reported result. Metric abbreviations and arrows as in Table~\ref{tab:table-a-uieb-lsui-256}.}\label{tab:table-b-other-datasets-256}
\small
\begin{tabular*}{\linewidth}{@{\extracolsep{\fill}}lcccccc}
\toprule
Method & PSNR$\uparrow$ & SSIM$\uparrow$ & LPIPS$\downarrow$ & FID$\downarrow$ & UCIQE$\uparrow$ & UIQM$\uparrow$ \\
\midrule
\multicolumn{7}{l}{\textit{UFO-120}} \\
UIE-WD & 19.4465 & 0.8122 & 0.3237 & 64.2594 & 0.5925 & \textbf{4.4257} \\
UWCNN & 20.7244 & 0.8625 & 0.2909 & 74.2507 & 0.5905 & 4.0840 \\
Water-Net & 24.2772 & 0.8780 & 0.2591 & 52.9071 & \underline{0.6010} & 4.3494 \\
U-shape & 29.5540 & \underline{0.9111} & 0.1792 & 38.8640 & 0.5997 & 4.2437 \\
WF-Diff & \textbf{30.6599} & 0.9044 & \underline{0.1336} & \underline{36.0713} & 0.5877 & 4.2971 \\
DiffWater & 22.6039 & 0.7992 & 0.3047 & 58.1146 & \textbf{0.6095} & \underline{4.3919} \\
Ours & \underline{30.3101} & \textbf{0.9146} & \textbf{0.1138} & \textbf{33.7577} & 0.5904 & 4.2741 \\
\midrule
\multicolumn{7}{l}{\textit{EUVP}} \\
UIE-WD & 19.3659 & 0.7946 & 0.3412 & 46.1214 & 0.5797 & \textbf{4.4398} \\
UWCNN & 21.5804 & 0.8521 & 0.3238 & 51.8359 & 0.5778 & 4.1469 \\
Water-Net & 24.4686 & 0.8644 & 0.2876 & 39.9792 & \textbf{0.5931} & \underline{4.4131} \\
U-shape & \textbf{31.4794} & \textbf{0.9030} & 0.1634 & \textbf{22.9462} & \underline{0.5886} & 4.2795 \\
WF-Diff & \underline{29.9486} & \underline{0.8945} & \underline{0.1589} & \underline{23.0850} & 0.5800 & 4.3459 \\
DiffWater & -- & -- & -- & -- & -- & -- \\
Ours & 29.2063 & 0.8931 & \textbf{0.1503} & 26.3554 & 0.5826 & 4.3222 \\
\midrule
 & \multicolumn{2}{c}{\textit{U45}} & \multicolumn{2}{c}{\textit{C60}} &  &  \\
\cmidrule(lr){2-3}\cmidrule(lr){4-5}
Method & UCIQE$\uparrow$ & UIQM$\uparrow$ & UCIQE$\uparrow$ & UIQM$\uparrow$ &  &  \\
\midrule
UIE-WD & 0.5647 & 3.9488 & 0.5227 & \underline{4.0000} &  &  \\
UWCNN & 0.5287 & 3.8195 & 0.5210 & 3.5638 &  &  \\
Water-Net & \underline{0.5755} & \textbf{4.2914} & \textbf{0.5637} & \textbf{4.0146} &  &  \\
U-shape & \textbf{0.5860} & \underline{4.2262} & \underline{0.5585} & 3.8208 &  &  \\
WF-Diff & -- & -- & -- & -- &  &  \\
DiffWater & -- & -- & -- & -- &  &  \\
Ours & 0.5626 & 4.0394 & 0.5501 & 3.7271 &  &  \\
\bottomrule
\end{tabular*}
\end{table}

\subsection*{Efficiency and single-step inference}

Table~\ref{tab:runtime-uieb-lsui} compares inference time on UIEB. WaveFlow-UIE is substantially faster than WF-Diff, reducing native-resolution runtime from $3392.32$ ms to $444.83$ ms and the $256\times256$ runtime from $301.30$ ms to $79.31$ ms. The diffusion-based DiffWater is far slower still, at about $38.4$\,s per image at $256\times256$. This supports the motivation for replacing the two-stage WF-Diff pipeline with a single flow-based pipeline. However, WaveFlow-UIE remains slower than lightweight feed-forward methods such as UIE-WD and Water-Net because it still performs multiple flow update steps during inference.

\begin{table}[!b]
\centering
\caption{\textbf{Inference time comparison on the public UIEB split (ms per image).} Lower is better. The Ours row uses WaveFlow-UIEB on UIEB. Native Time is the average of one timed pass per test image at its native resolution, without warm-up; 256 Time is the average of 100 timed passes on a $256\times256$ input after 20 warm-up passes. Because the two columns use different timing protocols, times should be compared only within a column. $^\dagger$DiffWater (2{,}000 sampling steps) was timed with one pass per test image at $256\times256$. Dashes mark combinations with no reported result.}\label{tab:runtime-uieb-lsui}
\small
\setlength{\tabcolsep}{12pt}
\begin{tabular}{lcc}
\toprule
Method & \multicolumn{2}{c}{UIEB} \\
\cmidrule(lr){2-3}
 & Native Time$\downarrow$ & 256 Time$\downarrow$ \\
\midrule
UIE-WD & 24.28 & 3.32 \\
UWCNN & 141.12 & 12.50 \\
Water-Net & 118.50 & 4.68 \\
U-shape & -- & 28.01 \\
WF-Diff & 3392.32 & 301.30 \\
DiffWater & -- & 38437.57$^\dagger$ \\
Ours & 444.83 & 79.31 \\
\bottomrule
\end{tabular}
\end{table}

\subsection*{Ablation studies}

Table~\ref{tab:ablation-uieb-256} reports the ablation study on UIEB at standardized $256\times256$ resolution. Its upper block evaluates the released WaveFlow-UIEB checkpoint with the default five-step inference and with a single flow step ($K=1$), and reports timing for both. Its lower block measures the contribution of individual components under a controlled protocol: the full model and four variants that remove the physics prior, the frequency weighting, the LPIPS loss, or the Lab color loss are each trained with identical settings (full-precision training with the same $100$k-iteration schedule, final checkpoint, seed 42, as in the follow-up comparison reported in the Discussion; see the controlled retraining protocol in Methods) and evaluated in the same way. Because a single training run varies with the random seed, the full model is also trained with two further seeds, and the lower block reports its mean and standard deviation over the three seeds.

\begin{table}[!t]
\centering
\caption{\textbf{UIEB ablation results at standardized 256$\times$256 resolution.} Upper block: the released WaveFlow-UIEB checkpoint; its times were re-measured with the protocol of the 256 Time column of Table~\ref{tab:runtime-uieb-lsui} (20 warm-up and 100 timed passes), so the full-model time differs slightly from that table. Lower block: every variant retrained under the same protocol (see Methods); the ablations use seed 42, and the full model is shown for seed 42 and as the mean and sample standard deviation (s.d., $n = 3$) over seeds 42, 1, and 2. PSNR is in dB. Dashes mark combinations with no reported result. Metric abbreviations and arrows as in Table~\ref{tab:table-a-uieb-lsui-256}.}\label{tab:ablation-uieb-256}
\small
\setlength{\tabcolsep}{3pt}
\begin{tabular*}{\linewidth}{@{\extracolsep{\fill}}lccccccc}
\toprule
Configuration & PSNR$\uparrow$ & SSIM$\uparrow$ & LPIPS$\downarrow$ & FID$\downarrow$ & UCIQE$\uparrow$ & UIQM$\uparrow$ & Time$\downarrow$ (ms) \\
\midrule
\multicolumn{8}{l}{\textit{Released WaveFlow-UIEB checkpoint}} \\
Full model ($K{=}5$) & 24.9818 & 0.9316 & 0.1390 & 29.7382 & 0.5835 & 4.0450 & 73.05 \\
$K{=}1$ single-step & 24.9382 & 0.9306 & 0.1403 & 30.6582 & 0.5827 & 4.0638 & 17.02 \\
\midrule
\multicolumn{8}{l}{\textit{Controlled retraining: 100k iterations, full precision, final checkpoint}} \\
Full model, mean (3 seeds) & 24.27 & 0.9256 & 0.1480 & 32.01 & 0.5779 & 4.039 & -- \\
Full model, s.d. (3 seeds) & 0.23 & 0.0020 & 0.0056 & 1.79 & 0.0029 & 0.031 & -- \\
Full model (seed 42) & 24.0039 & 0.9233 & 0.1541 & 33.9972 & 0.5811 & 4.0034 & -- \\
No physics prior & 24.0420 & 0.9241 & 0.1558 & 36.1105 & 0.5801 & 4.0742 & -- \\
No frequency weighting & 24.6787 & 0.9303 & 0.1385 & 31.4963 & 0.5801 & 4.0573 & -- \\
No LPIPS loss & 24.7934 & 0.9303 & 0.1394 & 30.8180 & 0.5868 & 4.0498 & -- \\
No Lab color loss & 24.3324 & 0.9301 & 0.1477 & 32.6765 & 0.5721 & 4.0956 & -- \\
\bottomrule
\end{tabular*}
\end{table}

The single-step setting is the most practically important result: reducing inference from five flow steps to one changes PSNR only from $24.98$ to $24.94$, SSIM only from $0.9316$ to $0.9306$, LPIPS only from $0.1390$ to $0.1403$, and FID only from $29.74$ to $30.66$, while reducing runtime from $73.05$ ms to $17.02$ ms, a $4.29\times$ speedup when both settings are timed under the same protocol. This shows that most of the restoration performance can be retained with a much cheaper inference process. The controlled ablation gives a less clear-cut picture than the design motivation suggests. Removing the physics prior leaves PSNR and SSIM within the range of the three full-model seeds but slightly worsens LPIPS ($0.1558$) and FID ($36.11$), so its benefit is limited to perceptual quality. Removing the Lab color loss keeps PSNR, LPIPS, and FID within the seed range and mainly lowers the UCIQE color score. Removing the frequency weighting or the LPIPS loss does not degrade the model: both variants reach higher PSNR and SSIM and lower LPIPS than all three full-model seeds. Each variant was trained with a single seed, so differences of the order of the full model's seed-to-seed standard deviation ($0.23$~dB PSNR) are not meaningful; nevertheless, the results suggest that, at this training budget, neither the high-frequency weighting nor the LPIPS term contributes to the reported quality, and that the gains come mainly from the wavelet-domain network itself.

\subsection*{Native-resolution results}

\subsubsection*{Native UIEB and LSUI}

Supplementary Table S1 evaluates the same datasets as Table~\ref{tab:table-a-uieb-lsui-256} at their native image resolutions. WaveFlow-UIE achieves the best PSNR, SSIM, LPIPS, and FID on UIEB, and the best PSNR, LPIPS, and FID on LSUI. This indicates that the proposed model is not only tuned to the resized $256\times256$ setting, but also generalizes well to native-resolution evaluation. The slightly lower UIQM values show that non-reference quality measures do not always align with full-reference and perceptual metrics, especially when color and contrast are enhanced differently by each method. U-shape is omitted because its released checkpoint is tied to $256\times256$ inputs.

\subsubsection*{Native UFO-120, EUVP, U45, and C60}

Supplementary Table S2 reports the native-resolution counterpart of Table~\ref{tab:table-b-other-datasets-256}. On UFO-120 and EUVP, WF-Diff obtains the best PSNR, SSIM, and FID, while WaveFlow-UIE obtains the best LPIPS and the second-best PSNR, SSIM, and FID. The EUVP rows match those in Table~\ref{tab:table-b-other-datasets-256} because the EUVP test images are already $256\times256$. On the reference-free U45 and C60 sets, WaveFlow-UIE does not lead the UCIQE or UIQM rankings, consistent with the $256\times256$ results. No native C60 result is reported for WF-Diff because the reproduced run exceeded the available GPU memory on the large C60 images.

\subsubsection*{Native ablation results}

Supplementary Table S3 repeats the ablation of Table~\ref{tab:ablation-uieb-256} at native resolution. The single-step variant ($K=1$) remains close to the full model (PSNR $24.71$ vs.~$24.68$~dB, SSIM $0.9344$ vs.~$0.9350$, LPIPS $0.1591$ vs.~$0.1589$, and FID $27.75$ vs.~$27.70$) while reducing the measured runtime from $444.83$~ms to $117.43$~ms. In the controlled block, removing the physics prior or the Lab color loss leaves PSNR and FID within the range of the three full-model seeds ($23.67$--$24.10$~dB PSNR), whereas removing the frequency weighting or the LPIPS loss again gives higher PSNR ($24.44$ and $24.38$~dB) and lower LPIPS and FID than all three seeds, in line with the $256\times256$ results.

\section*{Discussion}

WaveFlow-UIE is strongest on paired restoration benchmarks, where it consistently achieves strong perceptual quality and a favorable runtime-quality tradeoff relative to the released WF-Diff baseline. However, performance on fully reference-free challenge sets remains more mixed, and the model is still slower than lightweight feed-forward baselines because inference remains iterative. The current study also relies on a reproduced public-checkpoint benchmark rather than an exact reimplementation of every original paper protocol, so our claims should be interpreted within that unified local evaluation setting.

Follow-up experiments also clarify where the gains come from. Because the target velocity $v^\star = w_y - w_x$ does not depend on $\tau$, the single-step output $w_x + v_\theta(w_x, 0, w_x, p)$ has exactly the form of a direct wavelet-domain residual prediction. We therefore compared flow-matching training with training only at $\tau = 0$, i.e., direct residual regression, keeping everything else identical (network, losses, optimizer, data, full-precision training for a fixed $100$k iterations, and final checkpoints; three seeds per setting). On UIEB at $256\times256$, direct regression reached $26.11 \pm 0.31$~dB PSNR, compared with $24.27 \pm 0.23$~dB for flow matching with $K{=}5$ (mean $\pm$ s.d., $n = 3$ seeds; Supplementary Table S4), and every regression seed outperformed every flow seed on PSNR, SSIM, LPIPS, and FID. (The flow result is below the $24.98$~dB of Table~\ref{tab:ablation-uieb-256}, which comes from a longer run with checkpoint selection on the test split; see Implementation details in Methods.) The quality and speed reported in this paper should therefore be attributed to the single-pass wavelet-domain residual network and its training objective rather than to flow matching itself.

The physics branch should also be read as a learned conditioning input rather than as a calibrated estimate of the water medium, because nothing in the training objective ties $A$ to the true veiling light. On the 90 UIEB test images, the ambient light predicted by the released WaveFlow-UIEB checkpoint is close to zero in every channel (at most $0.0012$ on a scale from $0$ to $1$), and the largest value from any flow-matching checkpoint we examined is $0.125$ (Supplementary Table S5). In the flow models the ambient-light output therefore carries little information about color cast, and the conditioning comes mainly from the transmission map, which is consistent with the small effect of removing the branch in the controlled ablation. The direct-regression runs, in contrast, predict a blue-green ambient light (mean of about $(0.15, 0.45, 0.50)$ in RGB); we did not investigate the cause of this difference.

The method has potentially positive impact for underwater inspection, marine exploration, and robotic perception by improving visibility in degraded underwater imagery. At the same time, restoration errors may alter color or structure in ways that could affect downstream decisions if the outputs are used without validation in safety-critical settings. For that reason, the method should be treated as a perceptual restoration tool rather than a substitute for task-specific robustness evaluation.

\section*{Conclusions}

We presented WaveFlow-UIE, a single-pipeline wavelet-domain flow model for underwater image restoration. The method keeps the key frequency-domain insight of wavelet-based restoration while removing the need for a separate diffusion-style refinement stage. Across paired and cross-dataset benchmarks, WaveFlow-UIE shows a strong quality-efficiency tradeoff relative to the released WF-Diff baseline and particularly strong performance on perceptual metrics. The strongest practical result is that a single-step variant retains nearly all of the five-step model quality while delivering about a $4\times$ speedup on UIEB at $256\times256$ resolution. The strongest evidence in the current benchmark is on paired restoration tasks, while performance on fully reference-free challenge sets is more mixed and remains a useful direction for future work.

\section*{Data availability}


The datasets analyzed in this study are publicly available, including
\href{https://li-chongyi.github.io/proj_benchmark.html}{UIEB and C60},
\href{https://lintaopeng.github.io/_pages/UIE%20Project%20Page.html}{LSUI},
\href{https://irvlab.cs.umn.edu/resources/ufo-120-dataset}{UFO-120},
\href{https://irvlab.cs.umn.edu/resources/euvp-dataset}{EUVP}, and
\href{https://github.com/IPNUISTlegal/underwater-test-dataset-U45-}{U45}.
The numerical results generated in this study are reported in the tables and the Supplementary Information. Restored images and per-image metrics are available from the corresponding author on reasonable request.

\section*{Code availability}

The code used for training and evaluation, including the configurations and scripts of the controlled retraining, will be made public after publication of this article.

\section*{Acknowledgements}

The authors would like to acknowledge the Saudi Data and AI Authority (SDAIA) and King Fahd University of Petroleum \& Minerals (KFUPM) through the SDAIA-KFUPM Joint Research Center for Artificial Intelligence for providing computational resources.

\section*{Funding}

This work was supported by King Fahd University of Petroleum \& Minerals (KFUPM) under grant numbers EC241013, IN26117, INAI2605 and CAI02704.

\section*{Author contributions}

M.A.N. conceived the model. A.A. implemented the method, carried out the training and experiments, and wrote the manuscript. A.A., M.A.N., and H.A.N. analysed the results. M.B. supervised the project, secured the funding, and revised the manuscript. All authors reviewed the manuscript.

\section*{AI declaration}

Claude Code (Anthropic) was used by A.A. and M.A.N., and Codex (OpenAI) was used by A.A., to check code,
to analyze reported results, and to proofread the paper. 
The authors reviewed and verified all AI-assisted content and take full responsibility for the content of this article.

\section*{Additional information}

\noindent\textbf{Competing interests:} The authors declare no competing interests.

\noindent\textbf{Supplementary information} is available for this paper (Supplementary Tables S1 to S5).

\noindent\textbf{Correspondence} and requests for materials should be addressed to M.B.


\begin{thebibliography}{25}
\bibitem{jaffe1990} Jaffe, J. S. Computer modeling and the design of optimal underwater imaging systems. \emph{IEEE J. Ocean. Eng.} \textbf{15}, 101--111 (1990).

\bibitem{akkaynak2018} Akkaynak, D. \& Treibitz, T. A revised underwater image formation model. In \emph{Proc. IEEE/CVF Conference on Computer Vision and Pattern Recognition} 6723--6732 (2018).

\bibitem{islam2020fast} Islam, M. J., Xia, Y. \& Sattar, J. Fast underwater image enhancement for improved visual perception. \emph{IEEE Robot. Autom. Lett.} \textbf{5}, 3227--3234 (2020).

\bibitem{li2020uieb} Li, C. \emph{et al.} An underwater image enhancement benchmark dataset and beyond. \emph{IEEE Trans. Image Process.} \textbf{29}, 4376--4389 (2020).

\bibitem{li2020uwcnn} Li, C., Anwar, S. \& Porikli, F. Underwater scene prior inspired deep underwater image and video enhancement. \emph{Pattern Recognit.} \textbf{98}, 107038 (2020).

\bibitem{li2021ucolor} Li, C. \emph{et al.} Underwater image enhancement via medium transmission-guided multi-color space embedding. \emph{IEEE Trans. Image Process.} \textbf{30}, 4985--5000 (2021).

\bibitem{peng2023ushape} Peng, L., Zhu, C. \& Bian, L. U-shape transformer for underwater image enhancement. \emph{IEEE Trans. Image Process.} \textbf{32}, 3066--3079 (2023).

\bibitem{zhao2024wavelet} Zhao, C., Cai, W., Dong, C. \& Hu, C. Wavelet-based Fourier information interaction with frequency diffusion adjustment for underwater image restoration. In \emph{Proc. IEEE/CVF Conference on Computer Vision and Pattern Recognition} 8281--8291 (2024).

\bibitem{ho2020ddpm} Ho, J., Jain, A. \& Abbeel, P. Denoising diffusion probabilistic models. In \emph{Advances in Neural Information Processing Systems} Vol. 33, 6840--6851 (2020).

\bibitem{lipman2023flow} Lipman, Y., Chen, R. T. Q., Ben-Hamu, H., Nickel, M. \& Le, M. Flow matching for generative modeling. In \emph{Proc. International Conference on Learning Representations} (2023).

\bibitem{liu2023rectified} Liu, X., Gong, C. \& Liu, Q. Flow straight and fast: learning to generate and transfer data with rectified flow. In \emph{Proc. International Conference on Learning Representations} (2023).

\bibitem{he2011dcp} He, K., Sun, J. \& Tang, X. Single image haze removal using dark channel prior. \emph{IEEE Trans. Pattern Anal. Mach. Intell.} \textbf{33}, 2341--2353 (2011).

\bibitem{drews2013udcp} Drews, P. Jr, do Nascimento, E., Moraes, F., Botelho, S. \& Campos, M. Transmission estimation in underwater single images. In \emph{Proc. IEEE International Conference on Computer Vision Workshops} 825--830 (2013).

\bibitem{zhang2018lpips} Zhang, R., Isola, P., Efros, A. A., Shechtman, E. \& Wang, O. The unreasonable effectiveness of deep features as a perceptual metric. In \emph{Proc. IEEE/CVF Conference on Computer Vision and Pattern Recognition} 586--595 (2018).

\bibitem{peng2017ibla} Peng, Y.-T. \& Cosman, P. C. Underwater image restoration based on image blurriness and light absorption. \emph{IEEE Trans. Image Process.} \textbf{26}, 1579--1594 (2017).

\bibitem{ma2022uiewd} Ma, Z. \& Oh, C. A wavelet-based dual-stream network for underwater image enhancement. In \emph{Proc. IEEE International Conference on Acoustics, Speech and Signal Processing} 2769--2773 (2022).

\bibitem{wang2021uiec2net} Wang, Y., Guo, J., Gao, H. \& Yue, H. UIEC$^2$-Net: CNN-based underwater image enhancement using two color space. \emph{Signal Process. Image Commun.} \textbf{96}, 116250 (2021).

\bibitem{guan2024diffwater} Guan, M. \emph{et al.} DiffWater: underwater image enhancement based on conditional denoising diffusion probabilistic model. \emph{IEEE J. Sel. Top. Appl. Earth Obs. Remote Sens.} \textbf{17}, 2319--2335 (2024).

\bibitem{mallat1989} Mallat, S. G. A theory for multiresolution signal decomposition: the wavelet representation. \emph{IEEE Trans. Pattern Anal. Mach. Intell.} \textbf{11}, 674--693 (1989).

\bibitem{islam2020sesr} Islam, M. J., Luo, P. \& Sattar, J. Simultaneous enhancement and super-resolution of underwater imagery for improved visual perception. In \emph{Proc. Robotics: Science and Systems} (2020).

\bibitem{li2019fusion} Li, H., Li, J. \& Wang, W. A fusion adversarial underwater image enhancement network with a public test dataset. Preprint at \url{https://arxiv.org/abs/1906.06819} (2019).

\bibitem{wang2004ssim} Wang, Z., Bovik, A. C., Sheikh, H. R. \& Simoncelli, E. P. Image quality assessment: from error visibility to structural similarity. \emph{IEEE Trans. Image Process.} \textbf{13}, 600--612 (2004).

\bibitem{heusel2017gans} Heusel, M., Ramsauer, H., Unterthiner, T., Nessler, B. \& Hochreiter, S. GANs trained by a two time-scale update rule converge to a local Nash equilibrium. In \emph{Advances in Neural Information Processing Systems} Vol. 30 (2017).

\bibitem{yang2015uciqe} Yang, M. \& Sowmya, A. An underwater color image quality evaluation metric. \emph{IEEE Trans. Image Process.} \textbf{24}, 6062--6071 (2015).

\bibitem{panetta2015uiqm} Panetta, K., Gao, C. \& Agaian, S. Human-visual-system-inspired underwater image quality measures. \emph{IEEE J. Ocean. Eng.} \textbf{41}, 541--551 (2016).
\end{thebibliography}
\end{document}